\pdfoutput=1

\documentclass[11pt]{article}

\usepackage{EACL2023}

\usepackage{times}
\usepackage{latexsym}

\usepackage[T1]{fontenc}

\usepackage[utf8]{inputenc}

\usepackage{microtype}

\usepackage{inconsolata}

\title{Sparse Probability of Agreement}

\author{Jeppe N\o{}rregaard \\
  IT University of Denmark \\
  \texttt{jeno@itu.dk} \\\And
  Leon Derczynski \\
  IT University of Denmark \\
  \texttt{leod@itu.dk} \\}

\usepackage{amsmath}
\usepackage{amsfonts}
\usepackage{amssymb}
\usepackage{xspace}
\usepackage{multicol}
\usepackage{enumitem}
\usepackage{booktabs}
\usepackage{enumitem}
\usepackage{mathtools}

\usepackage{tikz}

\usepackage[textsize=scriptsize, textwidth=2.1cm, disable]{todonotes}		
\presetkeys{todonotes}{backgroundcolor=green, linecolor=black}{}

\newcommand{\veri}[1][$\checkmark$]{\todo[backgroundcolor=white,bordercolor=white,caption={}]{\small #1}}

\usepackage{array}			

\newcounter{rowcount}

\newcolumntype{L}[2] 	
{>{\raggedright\let\newline\\\arraybackslash\hspace{0pt}}#1{#2}}
\newcolumntype{C}[2]	
{>{\centering\let\newline\\\arraybackslash\hspace{0pt}}#1{#2}}
\newcolumntype{R}[2]	
{>{\raggedleft\let\newline\\\arraybackslash\hspace{0pt}}#1{#2}}

 \newcommand{\mathpath}{resources/}

\usepackage{amsmath,amsfonts,amssymb,bm,upgreek,cprotect}

\SetSymbolFont{largesymbols}{bold}{OMX}{txex}{b}{n}  

\ifthenelse{\isundefined{\mathpath}}{%
    \newcommand{\mathpath}{resources/}
}{%
}%

\let\r\right
\let\l\left

\newcommand{\appropto}{\mathrel{\vcenter{
	\offinterlineskip\halign{\hfil$##$\cr
		\propto\cr\noalign{\kern2pt}\sim\cr\noalign{\kern-2pt}}}}}

\renewcommand{\digamma}{\psi}                                   

\DeclareMathOperator{\E}{\mathbb{E}}                            
\DeclareMathOperator{\var}{var}                                 

\usepackage{accents}  
\usepackage{etoolbox}  

\makeatletter
\newcommand{\sbullet}{%
  \hbox{\fontfamily{lmr}\fontsize{.4\dimexpr(\f@size pt)}{0}\selectfont\textbullet}}

\makeatother

\DeclareMathAlphabet{\mathsfit}{\encodingdefault}{\sfdefault}{m}{sl}
\SetMathAlphabet{\mathsfit}{bold}{\encodingdefault}{\sfdefault}{bx}{n}

\def\vk{{\bf{k}}}

\def\vn{{\bf{n}}}

\def\vone{{\bm{1}}}

\def\evk{\text{k}}

\def\evn{\text{n}}

\newcommand{\I}{\mathbb{I}}

\newcommand{\breakcolumn}{\vfill\eject}

\newcommand{\varw}{\texttt{inv\_var}\xspace}
\newcommand{\varwclass}{$\texttt{inv\_var}_{\texttt{class}}$\xspace}

\newcommand{\spa}{\bar{P}_{\text{spa}}}

\begin{document}
\maketitle
\begin{abstract}
	Measuring inter-annotator agreement is important for annotation tasks, but many metrics require a fully-annotated set of data, where all annotators annotate all samples. 
	We define Sparse Probability of Agreement, SPA, which estimates the probability of agreement when not all annotator-item-pairs are available.
	We show that under certain conditions, SPA is an unbiased estimator, and we provide multiple weighing schemes for handling data with various degrees of annotation.
\end{abstract}

%

\section{Measuring Agreement}

Inter-annotator agreement (IAA) is the degree of agreement between independent annotators performing some task. High IAA scores indicate agreement between annotators. 

Commonly-used IAAs require a fully-annotated dataset (or subset), where all annotators annotate all instances. This can be both expensive and difficult to orchestrate. An alternative is to have all annotators annotate a subset of the dataset and measure agreement over this subset, though here the result will necessarily be  biased. We present a method for computing a sparse measure of agreement over the whole dataset, to alleviate this annotation-expense problem. \\

One simple measure of IAA is joint probability of agreement (PA), which is the probability any two annotators agree on a random item. If we compute the probability that two annotators agree on item $i$ the item/sample-agreement of item $i$ by:
\begin{align}
	P_i &= \frac{1}{n(n-1)} \sum_c^C n_{ic}(n_{ic} - 1).  \label{eq:item_agreement},
\end{align} 

then joint probability of agreement (PA) is the sample mean of the item-agreements:
\begin{align}
	\bar{P} &= \frac{1}{I} \sum_i^I P_i, \label{eq:mean_sample_agreement},
\end{align}
where $n$ is the number of annotations, $C$ the number of classes, $I$ the number of items, and $n_{ic}$ the number of annotations of item $i$ into class $c$.

PA is readily interpretable, and is the basis for many other measures, but is not always ideal for measuring IAA, because it does not take agreement-by-chance into account. That is, if annotators randomly select classes and the class distribution is skewed, there will be a high PA, despite the random guessing. \\

Alternatives for assessing agreement between annotators that takes randomness into account is therefore to use the kappa/alpha family of measures \cite[p. 299]{ide_handbook_2017}. 
A commonly used method is Fleiss' kappa \cite{fleiss_measuring_1971}, which is defined as:
\begin{align}
	\kappa = &\frac{\bar{P} - \bar{P}_e}{1 - \bar{P}_e}, \\
    \bar{P}_e = \sum_c p_c^2 &\qquad
    p_c = \frac{1}{nI} \sum_i n_{ic},
\end{align}
where $\bar{P}_e$ is the expected agreement-by-chance, and $p_c$ is the empirical class distribution. \\
Computing PA and its derivatives requires a fully annotated set of data, where all annotators have labelled all instances. In many cases this is not possible (e.g. most crowdsourced labelings).
This paper investigates computing PA sparsely, with missing/unfinished annotations. \\

\section{Background}

\subsection{Krippendorff's alpha}

A notable method from the kappa/alpha family is Krippendorff's alpha \cite{krippendorff_content_1980}, which uses the observed and expected \textit{disagreement} of annotations. The observed disagreement is:
\begin{align}
	D_o = \frac{1}{U} \sum_{i} \frac{n_i}{P(n_i, 2)} \sum_c \sum_k \delta(c, k) x_{cki},
\end{align}
where $n_i$ is the number of annotations of item $i$, $x_{nki}$ is the number of $(c, k)$ pairs for item $i$, $P$ is the permutation function: $P(n_i, 2) = n_i (n_i - 1)$, $\delta$ is a chosen difference metric and $U$ is the total number of pairable elements. Alpha is computed by $\alpha = 1 - \frac{D_o}{D_e}$, which is chance-corrected in the same way that the kappa-family, if we define disagreement as $D_o = 1 - \bar{P}$ and $D_e = 1 - \bar{P}_e$. $\alpha = 1$ at perfect agreement (zero disagreement). 
Krippendorff's alpha can handle missing data, as well as labelling that is nominal, ordinal, interval, ratio and more. \\

For Krippendorff's alpha, $\delta$ must be a difference metric, but if we violate this constraint and set $\delta(c, k) = \I(c \not= k)$ we have:
\begin{align}
    \bar{P} &= 1 - D_o \label{eq:krippendorf_like_siaa} \\
    &= 1 - \frac{1}{U} \sum_{i} \frac{n_i}{P(n_i, 2)} \sum_c \sum_k \I(c \not= k) x_{cki} \nonumber \\
    &= 1 - \frac{1}{U} \sum_{i} \frac{n_i}{n_i (n_i - 1)} \sum_c \sum_{k \not= c} x_{cki} \nonumber \\
    &= \frac{1}{U} \sum_{i} \frac{n_i}{n_i (n_i - 1)} \sum_c x_{cci} \nonumber \\
    &= \frac{1}{U} \sum_{i} n_i\frac{1}{n_i (n_i - 1)} \sum_c n_{ic} (n_{ic} - 1) \nonumber \\
    &= \frac{1}{U} \sum_{i} n_i P_i, \nonumber
\end{align}

\subsection{Missing Data}
\citet{van_oest_weighting_2021} propose an approach to the problem of missing data by generalizing chance-corrected measures to a Bayesian model, which can handle both missing data and weighing of error-types (similar to Krippendorff's alpha). \citet{de_raadt_kappa_2019} present three methods for handling missing data when computing Cohen's kappa \cite{cohen_coefficient_1960} (which only works for two classes). They use four methods for computing kappa with missing data: ignoring samples with a missing label; computing agreement with samples with both labels; using all labels for computing class distribution and expected agreement; and considering missing labels as a separate category (expanding Cohen's to Fleiss' kappa).

Fleiss' kappa is a generalisation of Cohen's kappa, but is not suitable for use in scenarios where individual annotators only annotate a subset of the data. IAA measures computed using Fleiss' kappa will have had to either subsample data to the set of instances that all annotators have seen, or ``re-use'' annotator ``slots'' for multiple annotators, or otherwise re-cast the annotation results. Precisely how this adaptation of annotations is implemented is not always clear from papers using the metric.

\section{Sparse Probability of Agreement}

Sparse Probability of Agreement (SPA) relaxes the constraint that all annotators label all instances. It is defined as:
\begin{align}
    \spa &= \frac{1}{\vone^{\top} \vk} \sum_i \evk_i \hat{P}_i \label{eq:spa_definition} \\
    \hat{P}_i &= \frac{1}{n_i (n_i - 1)} \sum_c^C n_{ic} (n_{ic} - 1), \nonumber
\end{align}
where crucially we have different numbers of annotations $n_i$ for each item $i$, and a weighing $\vk$ of the items. 
SPA remains interpretable as the probability of two random annotators agreeing on a random item, taken from the sets of annotators and items in the dataset.\\
SPA is a weighed micro-average of the annotation agreements of each item, and $\spa = \bar{P}$ when the same annotators annotate all items and $\vk = \vone$. Also, when $\vk = \vn$, $\spa$ matches the expression in \ref{eq:krippendorf_like_siaa}, which is the the agreement found, when violating the definition of Krippendorf's alpha, using $\delta(c, k) = \I(c \not= k)$. \\
%
%
%

As we will show, SPA is an unbiased estimator of PA.
SPA can monitor the inter-annotator agreement during the annotation process, allowing for intervening and improving task if the agreement does not meet expectations. It also solves common problems with crowd-sourcing annotations, where one cannot reliably ensure that all annotators finish all tasks.


\begin{figure}
	\centering
	\includegraphics[width=0.6\linewidth]{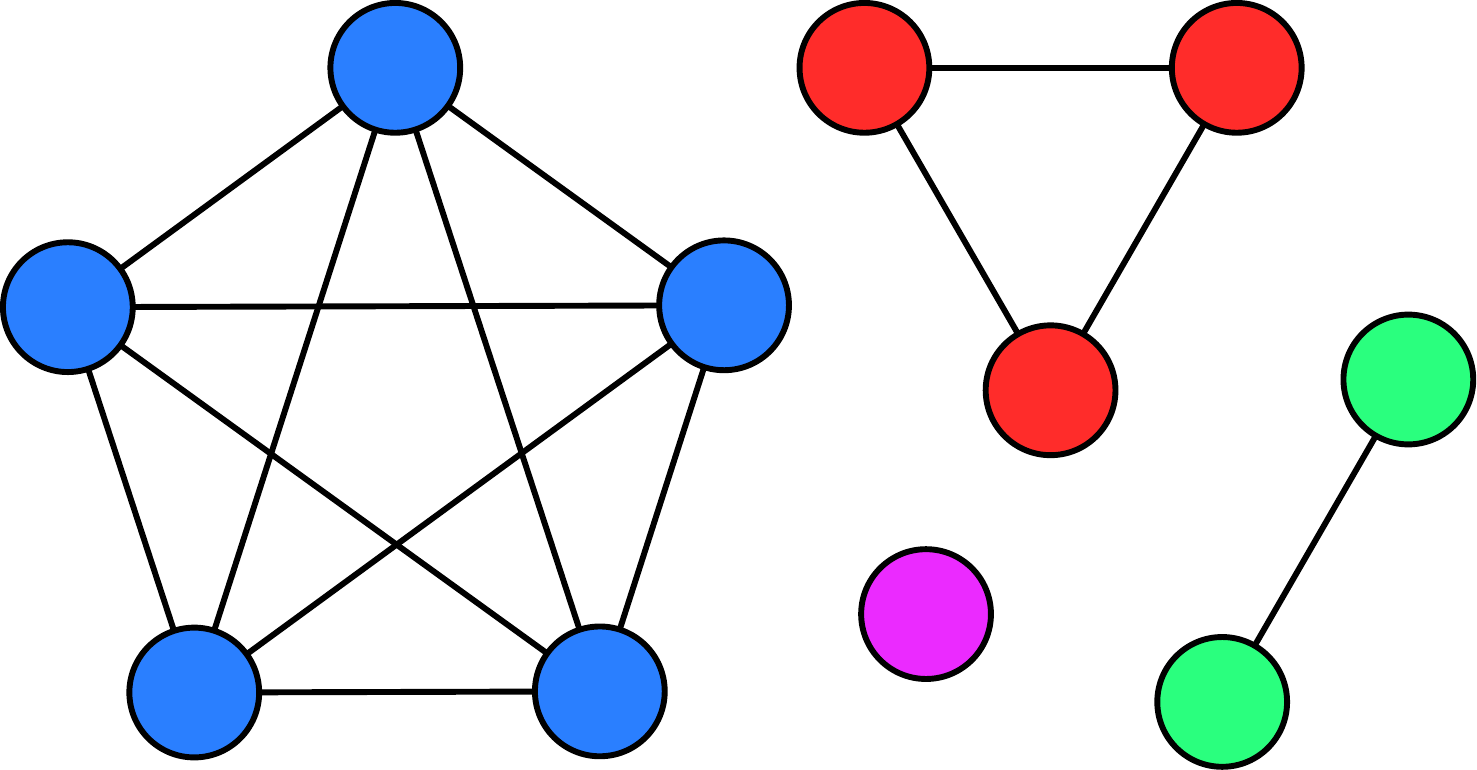}
	\caption{Example of annotation with 11 annotators. Five annotators agree on the blue category, three on the red category, two on the green category and one on the pink category. The number of edges is 14, while the number of possible edges is $\frac{11 (11 - 1)}{2} = 55$. The agreement is $\frac{14}{55} \approx 25.5\%$.} \label{fig:annotation_graph_example}
\end{figure}

\section{Properties of Sparse Agreement}

\subsection{Annotation of One Item}

Consider a single item, which has been annotated by $n$ annotators. We can consider the annotations an undirected graph, in the following way (example in Figure~\ref{fig:annotation_graph_example}).
Each annotation is a node, whose colour is the category. There are edges between all nodes of the same colour (all annotations that agree), but no edges in-between colour groups. The number of edges is:

\begin{align}
	n_{\text{edges}} 
		&= \sum_c \frac{n_c (n_c - 1)}{2} 
		= \frac{1}{2} \sum_c n_c (n_c - 1), \nonumber
\end{align}
where $c$ is a category/colour.

The total possible number of edges in a graph is:

\begin{align}
	N = \frac{n (n - 1)}{2},
\end{align}
which will be equal to $n_{\text{edges}}$ if all annotators agree. 

The agreement of the annotation is the number of edges over the total possible number of edges:
\begin{align}
	P &= \frac{n_{\text{edges}} }{N}
		= \frac{2}{n (n - 1)} \times \frac{1}{2} \sum_c n_c (n_c - 1) \nonumber \\
		&= \frac{1}{n (n - 1)} \sum_c n_c (n_c - 1),
\end{align}

which matches the expression in (\ref{eq:item_agreement}). Finally:

\begin{align}
	n_{\text{edges}} &= P \times N \label{eq:agreement_graph_formulation}.
\end{align}

\subsubsection{Removing One Annotation}

\begin{figure}
	\centering
	\includegraphics[width=0.7\linewidth]{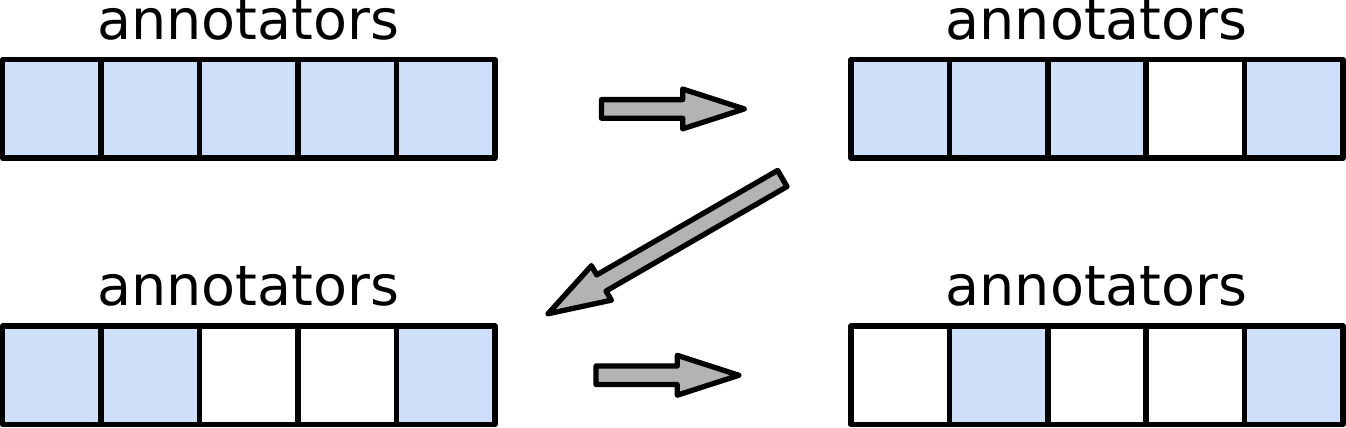}
	\caption{Blue cells indicate a known annotation, while while cells indicate a missing annotation. Initially we have an item which has been annotated by all 5 annotators (top-left corner). We can then randomly remove one annotation at a time to create a sparsely annotated item. The expected agreement remains the same.}
	\label{fig:sparse_annotation_single_item}
\end{figure}

Say we randomly remove one annotation. The new number of possible edges becomes:

\begin{align}
	N' = N - (n-1) = N - n + 1.
\end{align}

The expected degree (number of edges) of a random node is:

\begin{align}
	\E_j[ \deg(j) ] = P \times (n - 1),
\end{align}
and the expected new number of edges in the graph is therefore:

\begin{align}
	\E[ n'_{\text{edges}} ]
		&= n_{\text{edges}} - \E_j[ \deg(j) ]
\\
		&= n_{\text{edges}} - P \times (n - 1) \nonumber \\
		&= P \times N - P \times (n - 1) \nonumber
\\
		&= P (N - n + 1),
\nonumber
\end{align}
using (\ref{eq:agreement_graph_formulation}). 
The expected new agreement $P'$ is:

\begin{align}
	\E[P'] 
		&= \frac{ \E[n'_{\text{edges}}] }{ N' }
\label{eq:expected_agreement_on_item_equals_original}
		= \frac{ P (N - n + 1) }{ N - n + 1 }
\\
		&= P \; \frac{ N - n + 1 }{ N - n + 1 }
		= P. \nonumber
\end{align}

Therefore, when we randomly remove an annotation, the expected agreement remains the same. 

As exemplified in Figure~\ref{fig:sparse_annotation_single_item}, we can keep applying this trick going from $n$ annotations down to 2. The agreement will vary depending on which nodes we randomly select, but in expectation, the agreement will remain the same.

Note the two special cases:
\begin{align}
	n&=2, \qquad N=1, \\
		&\quad \E[P'] = P \; \frac{ 1 - 2 + 1 }{ 1 - 2 + 1 } = P \; \frac{0}{0} = \text{undef.} \nonumber\\[3mm]
	n&=1, \qquad N=0, \\
		&\quad \E[P'] = P \; \frac{ 0 - 1 + 1 }{ 0 - 1 + 1 } = P \; \frac{0}{0} = \text{undef.}
\nonumber
\end{align}
which make intuitive sense, as we cannot compute agreement with zero or one annotations.

\begin{figure}
	\centering
	\includegraphics[width=\linewidth]{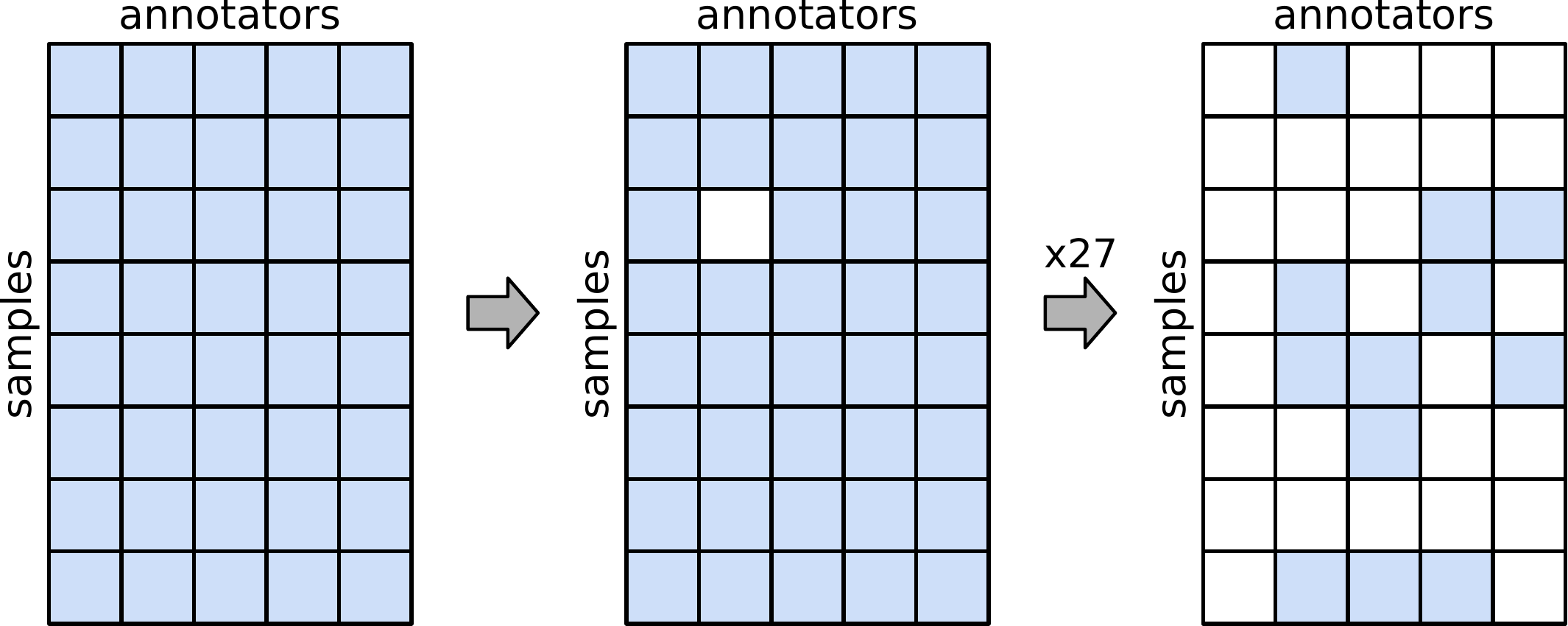}
	\caption{Blue cells indicate a known annotation, while while cells indicate a missing annotation. Initially we a (hypothetical) fully annotated dataset. We then remove a single annotation to exemplify. After removing 27 more annotation we end up with the sparse annotation dataset on the right. In the sparse annotation dataset 4 samples have too few annotations (0 or 1) to be used for computing inter-annotator agreement, while the remaining 4 have enough. $\spa$ of the sparse annotation matrix has the same expectation as the inter-annotator agreement of the full one.}
	\label{fig:sparse_annotation}
\end{figure}

\subsection{Multiple Items}

We now consider multiple items by using the SPA formulation from (\ref{eq:spa_definition}).
Note that $\spa$ is an unbiased estimator of $P$, as long as the weights $\vk$ are statistically independent of the item-agreements: $k_j \bot P_j$.

\subsubsection{Removing One Annotation}  \label{sec:multiple_removing_one_annotation}
Now say we randomly remove one annotation from item $i$. The expected mean sample-agreement is:

\begin{align}
	\E[\bar{P}] 
		&= \E\left[ \frac{1}{K} \left( \sum_{j \not= i}^N \evk_j P_j + \evk_i P'_i \right) \right]
\end{align}
\begin{align}
	\E[\bar{P}] 
		&= \frac{1}{K} \left( \sum_{j \not= i}^N \evk_j P_j + \evk_i \E[P'_i] \right) \nonumber \\
		&= \frac{1}{K} \left( \sum_{j \not= i}^N \evk_j P_j + \evk_i P_i \right) \nonumber \\
		&= \frac{1}{K} \sum_{j}^N \evk_j P_j = \bar{P}, 
\end{align}

using (\ref{eq:expected_agreement_on_item_equals_original}).

Computing the expected agreement on a dataset with missing annotations is therefore an unbiased estimator of the agreement of a hypothetically fully-annotated dataset.

\subsubsection{Removing Multiple Annotations}

We make two observations for randomly removing multiple annotations. First; we can repeatedly remove a single annotation like in Section~\ref{sec:multiple_removing_one_annotation} and the expectation $E[\bar{P}]$ will remain the same. Such a sparse annotation dataset is exemplified in figure \ref{fig:sparse_annotation}.
Second; if our removal is random, then basing the weights $\vk$ on the number of annotations of each items will satisfy the constraint $k_j \bot P_j$, keeping $E[\bar{P}]$ an unbiased estimator. In the following sections we discuss four intuitive and simple weighing schemes for $\vk$, and derive two more complicated ones, that are based on the variance of item-agreements.

\subsection{SPA Assumptions}
SPA makes one key assumption: \textit{The degree to which labels are absent must be independent of the true item-agreements $n_i \bot P_i$. }

For example, if items are randomly distributed to annotators and a random set of annotators do not finish some of their work, then SPA will work fine. On the other hand, if some samples are more likely to have missing labels (for example because they are more difficult to get), then we break the assumption on missingness. As distributing samples randomly between annotators is a very common practise, SPA is highly applicable.

SPA do not assume anything about the underlying data, labelling process, label distribution or noise structure of labels. Where some works that can handle missing data assume there is a "correct" class (for example \citet{van_oest_weighting_2021}). SPA does not make this assumption, but simply estimates the agreement of a specific dataset with a specific set of annotators, given randomly missing labels. This makes it useful for a broad range of cases. For example, in a survey asking people what their favourite food is, there is no correct class. But we can still use SPA to discuss agreement.

\subsection{Chance Correction}

Chance correcting SPA is harder as the naive approach (based on Fleiss kappa) becomes a biased estimator, due to two properties:

\begin{enumerate}
	\item Jensen's inequality \cite{jensen_sur_1906} notes that
		\begin{align}
			\frac{x}{\E\l[ 1 - \bar{P}_e\r]} \leq \E\l[ \frac{x}{1 - \bar{P}_e} \r]
		\end{align}
		which estimates using a sampled $\bar{P}_e$ a \textit{ratio estimator}; this estimator is biased; 
	\item $\bar{P}$ and $\bar{P}_e$ are statistically dependent, which makes the the numerator and denominator statistically dependent. This further makes the estimator biased.
\end{enumerate}

The second bias is also found in the standard way of computing Fleiss' kappa for fully annotated datasets, as $\bar{P}$ and $\bar{P}_e$ are most often computed on the same dataset. Also, ratio estimators have bias on the order of $O(n^{-1})$, which makes the estimator approximately unbiased for large sample sizes. An investigation of the chance-correction of SPA would be useful future work.

\section{Weighing Schemes}

We investigate 6 weighing schemes $\vk$ for computing SPA.

\subsection{Simple Weighing Schemes}

Four simple and intuitive weighing schemes are
\begin{description}
	\item[flat] All samples have weight 1. Inter-annotator agreement is a simple mean of agreement on samples.
	\item[annotations] Samples are weighed by the number of annotations (scales linearly with number of annotations), similarly to that of Krippendorff's alpha. 
	\item[annotations\_m1] Samples are weighed by the number of annotations minus 1. It scales linearly in the number of annotations and naturally assigns weight 0 to samples with 1 annotation.
	\item[edge] Samples are weighed by the number of edges. Weight scales quadratically with number of annotations and naturally assigns weight 0 to samples with 1 annotation.
\end{description}

\subsection{Inverse-Variance Weighting}
The weighing $\vk$ will not influence the expectation of the estimate, but it can influence the variance of the estimate. 
We naturally wish to select a weighing $\vk$ that minimizes this variance, which can be found using inverse-variance weighting, so that $\evk_j = \var[P_j]^{-1}$.
We therefore wish to estimate the variance of each item-agreement $\var[P_j]$.

\subsection{Expected Variance wo. Class Distribution}  \label{sec:exp_var_no_class_distribution}

If we have no knowledge about the class distribution, then the expected variance across all possible annotations, for $n$ annotators on $C$ classes, is:

\begin{align}
	\var[&\hat{P}]
	= \frac{1}{N^2 C^n 4} \bigg(\sum_{n_0=0}^n \sum_{n_1=0}^{\substack{\min \\ n_0, n - n_0}}  \label{eq:exp_var_no_class_distribution} \\
		& K_{csp}(n_0, n_1) \; n_0 n_1(n_0 - 1)(n_1 - 1) \nonumber \\
		&+ \sum_{n_c=0}^n K_{sps}(n_c) \; n_c^2 (n_c - 1)^2 \bigg) \nonumber \\
		&- \E[\hat{P}]^2, \nonumber
\end{align}

using the utility functions:

\begin{align}
	&K_{sps}(n_c) = {n \choose n_c} C (C - 1)^{n-n_c} \\ 
	&K_{csp}(n_0, n_1) \nonumber \\
		&= 2^{\I(n_0 = n_1)} {n \choose n_0} {n - n_0 \choose n_1} \nonumber \\
		&\qquad C (C - 1) (C - 2)^{n - n_0 - n_1},\nonumber
\end{align}

This can be computed in $O(n^2)$ time. The derivation is in Appendix \ref{apdx:item_variance_no_class_distribution}.

We denote the inverse-variance weights using this method \varw.

\subsection{Expected Variance w. Class Distribution}  \label{sec:exp_var_with_class_distribution}

If we know the class distribution (or perhaps can estimate it), the expected variance across all possible annotations, for $n$ annotators on $C$ classes with probability $p_c$ of class $c$, is:

\begin{align}
	&\var[\hat{P}]  \label{eq:exp_var_with_class_distribution}
	= 
		\frac{1}{4} 
		\sum_{\substack{c}} \sum_{\substack{c' \\ c \not= c'}} 
		\sum_{\evn_c}^n \sum_{\evn_{c'}}^{n - \evn_c}p(\evn_c, \evn_{c'}) \\ 
					&\qquad \evn_c \evn_{c'}(\evn_c - 1)(\evn_{c'} - 1) \nonumber \\
		&+ \frac{1}{4} \sum_{c} \sum_{\evn_c}^n p_c(\evn_c) \evn_c^2 (\evn_c - 1)^2 - \E[\hat{P}]^2, \nonumber
\nonumber
\end{align}

using the probabilities:

\begin{align}
	p(\evn_c) &= {n \choose \evn_c} p_c^{\evn_c} (1 - p_c)^{n - \evn_c} \\
	p(\evn_c, \evn_{c'})
		&= {n \choose \evn_c} {n - \evn_c \choose \evn_{c'}} \nonumber \\ 
		&\qquad p_c^{\evn_c} p_{c'}^{\evn_{c'}} (1 - p_c - p_{c'})^{n - \evn_c - \evn_{c'}}. \nonumber
\end{align}

This can be computed in $O(n^2C^2)$ time. The derivation is in Appendix \ref{apdx:item_variance_w_class_distribution}. In appendix \ref{append:var_with_class_distribution_uniform_special_case} we show that (\ref{eq:exp_var_no_class_distribution}) is the maximum-entropy special case of (\ref{eq:exp_var_with_class_distribution}), when the class distribution is uniform.
We denote the inverse-variance weights using this method \varwclass.

\subsection{One Annotation Case}

Due to the expected agreement term $E[P]$ in the computation of variances, the variances becomes undefined when only a single annotation for an item is provided. We set the variance of items with a single annotation to $\infty$, as this will set the inverse-variance weight to 0 for items with a single annotation.

\section{Experiments}

\subsection{Comparing Weighing Schemes}

\begin{figure}[t]
	\includegraphics[width=\linewidth]{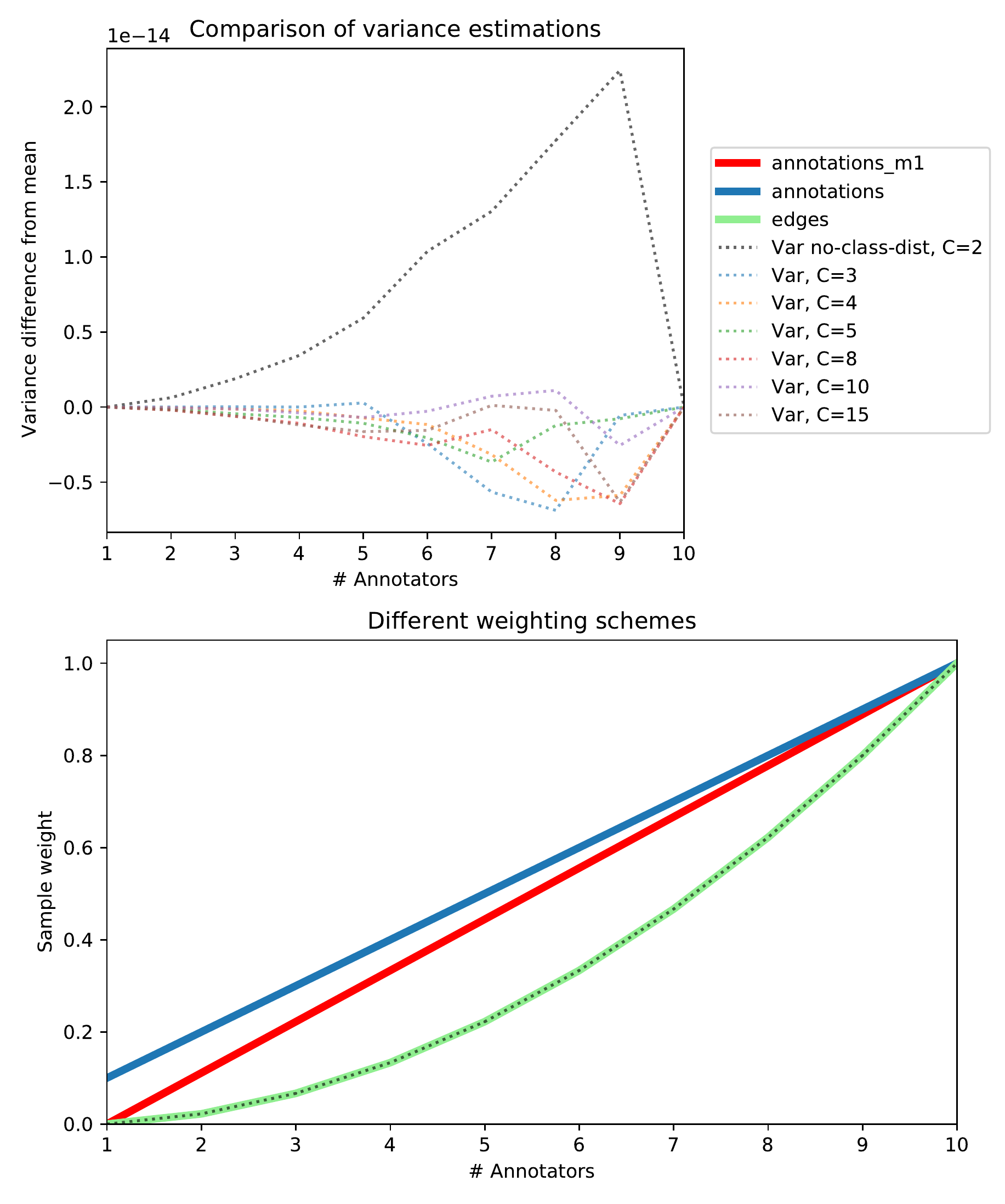}
	\caption{Weights of samples depending on number of annotations, for different weighing schemes. Top plot shows the weights for weighing schemes \texttt{annotations\_m1}, \texttt{annotations}, \texttt{edges} and using inverse-variance with no class-distribution for 2-classes. Weights are normalized to map the largest weight to 1. \newline
	The bottom plot shows difference in inverse-variance with no class-distributions for different numbers of classes.}
	\label{fig:weight_curves}
\end{figure}
\begin{figure}[t]
	\includegraphics[width=\linewidth]{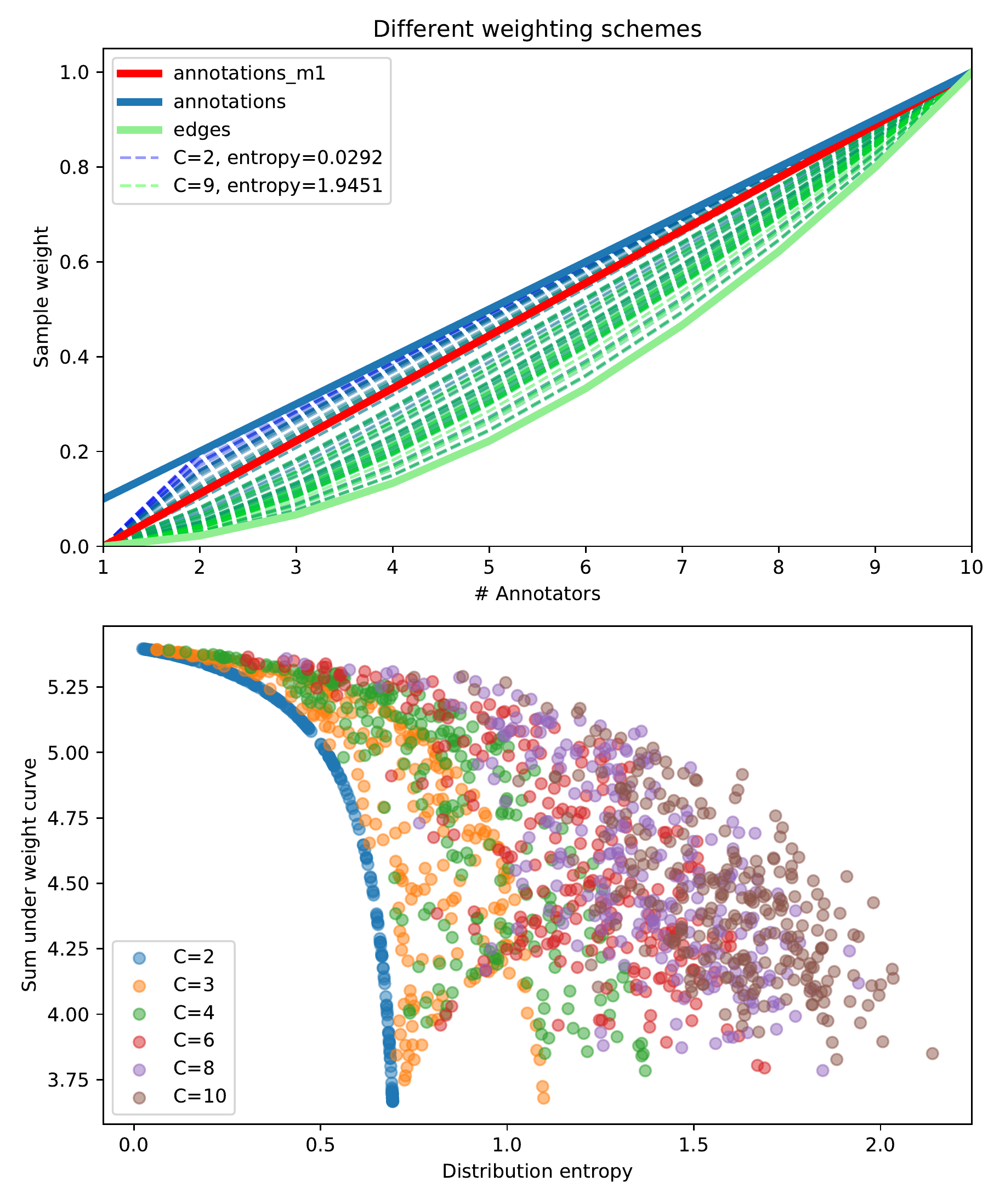}
	\caption{Comparison of inverse-variance weighing using class-distribution. Top plot shows the three simple weighing schemes and the computed inverse-variance weights using a set of randomly sampled class-distributions. We have coloured the curves according to the entropy of the distributions, so that high-entropy classes have a green colour and blue otherwise. \newline 
	The bottom plot shows a scatter plot of the area under the weight-curve over the entropy of the class-distribution.}
	\label{fig:weight_curves_class_dist}
\end{figure}

We investigate how the number of classes affect the \varw weights. We compute the (normalized) weights for items with 2 to 10 annotations, when the number of classes is 2-7. In figure \ref{fig:weight_curves} we plot these curves after subtracting the mean (for easy comparison). 
The \varw weights seems to be constant with respect to the number of classes (difference is so small it could be floating point errors). We have not been able to show theoretically why this is the case. 

In the bottom of Figure \ref{fig:weight_curves} we show the (normalized) weight curves of \texttt{annotations}, \texttt{annotations\_m1}, \texttt{edges} and \varw with $C=2$. All weighing schemes apply a lower weight to samples with few annotations, as expected. We also notice that \varw is almost identical to \texttt{edges}. 

For analysing the \varwclass weights, we randomly sample distributions, by uniformly sampling logits in the range $(-2, 4)$ and applying the softmax function to produce a distribution, for classes $C \in [2, 10]$. We sample 10 distributions for each $C$ and plot their weight curves (dashed lines) in the top plot of figure \ref{fig:weight_curves_class_dist}. We colour the lines depending on the distributions entropy, so that high-entropy lines are green and low entropy-lines are blue. It appears that \varwclass selects a weight-curve ranging from number of \texttt{annotations} to number of edges \texttt{edges}, depending on the class distribution, and that this correlates somewhat with the distribution's entropy. As previously noted, for the maximum-entropy distribution (a uniform distribution), we have $\text{\varwclass} = \text{\varw}$, which aligns with \texttt{edges}. 

At the bottom of Figure~\ref{fig:weight_curves_class_dist} we plot the sum-under-weight-curve (the sum of the weights for annotations 1-10) over the distributions entropy, for 250 sampled distributions for each $C$. We note that there is some relationship between the sum-under-weight-curve and entropy, but they do not directly correspond.

\begin{figure}[t]
	\includegraphics[width=0.91\linewidth]{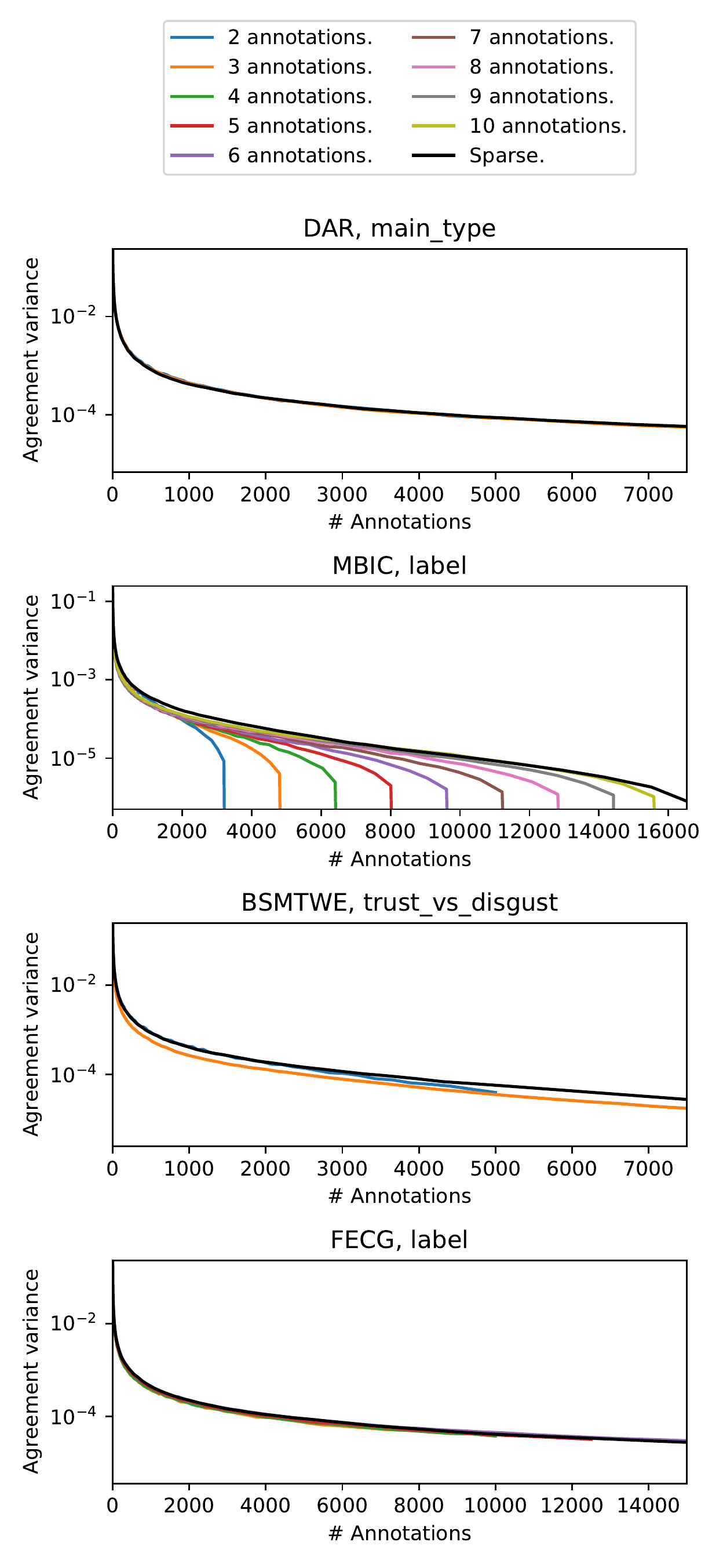}
	\caption{Adding annotations to inter-annotator dataset. As more annotations are added, the variance in resulting inter-annotator agreement decreases.} 
	\label{fig:subsampling_annotations_sparse_comparison}
\end{figure}

\begin{table*}[t]
    \centering
    \small
    \begin{tabular}{lrrrrrr}
\toprule
Scale: $10^2$ &    edge &   flat &  annotations &  annotations\_m1 &     var &  var\_p\_class \\
\midrule
MBIC, label                 & -0.4681 & 0.0000 &      -0.8458 &           \textbf{-1.0929} & -0.4681 &      \underline{-0.8676} \\
MBIC, factual               & -0.7894 & 0.0000 &      -0.9540 &           \textbf{-1.2747} & -0.7894 &      \underline{-1.0346} \\
BSMTWE, trust\_vs\_disgust    & -0.4333 & 0.0000 &      -0.3567 &           \underline{-0.5141} & -0.4333 &      \textbf{-0.5359} \\
BSMTWE, surprise\_vs\_antecip & -0.3322 & 0.0000 &      -0.2909 &           \underline{-0.4148} & -0.3322 &      \textbf{-0.4437} \\
BSMTWE, joy\_vs\_sadness      & -0.2194 & 0.0000 &      -0.2943 &           \textbf{-0.3966} & -0.2194 &       \underline{-0.3899} \\
BSMTWE, anger\_vs\_fear       & -0.1614 & 0.0000 &      -0.3129 &            \underline{-0.4089} & -0.1614 &      \textbf{-0.4118} \\
FECG, label                 &  0.2910 & \underline{0.0000} &      \textbf{-0.0444} &            0.0011 &  0.2910 &       0.2003 \\
DAR, main\_type              &  0.2019 & \textbf{0.0000} &       \underline{0.0089} &            0.0493 &  0.2019 &       0.1324 \\
\bottomrule
\end{tabular}

    \caption{Sum-of-curve from Figure~\ref{fig:weighing_subtract_mean} (and the full one in the appendix Figure \ref{fig:weighing_subtract_mean_full}), which compares the weighing schemes with using flat-weights (the baseline). Lower is better. Best performance is bold, second best is underline.}
    \label{tab:weights_comparison_table}
\end{table*}

\begin{figure}[t]
	\includegraphics[width=0.91\linewidth]{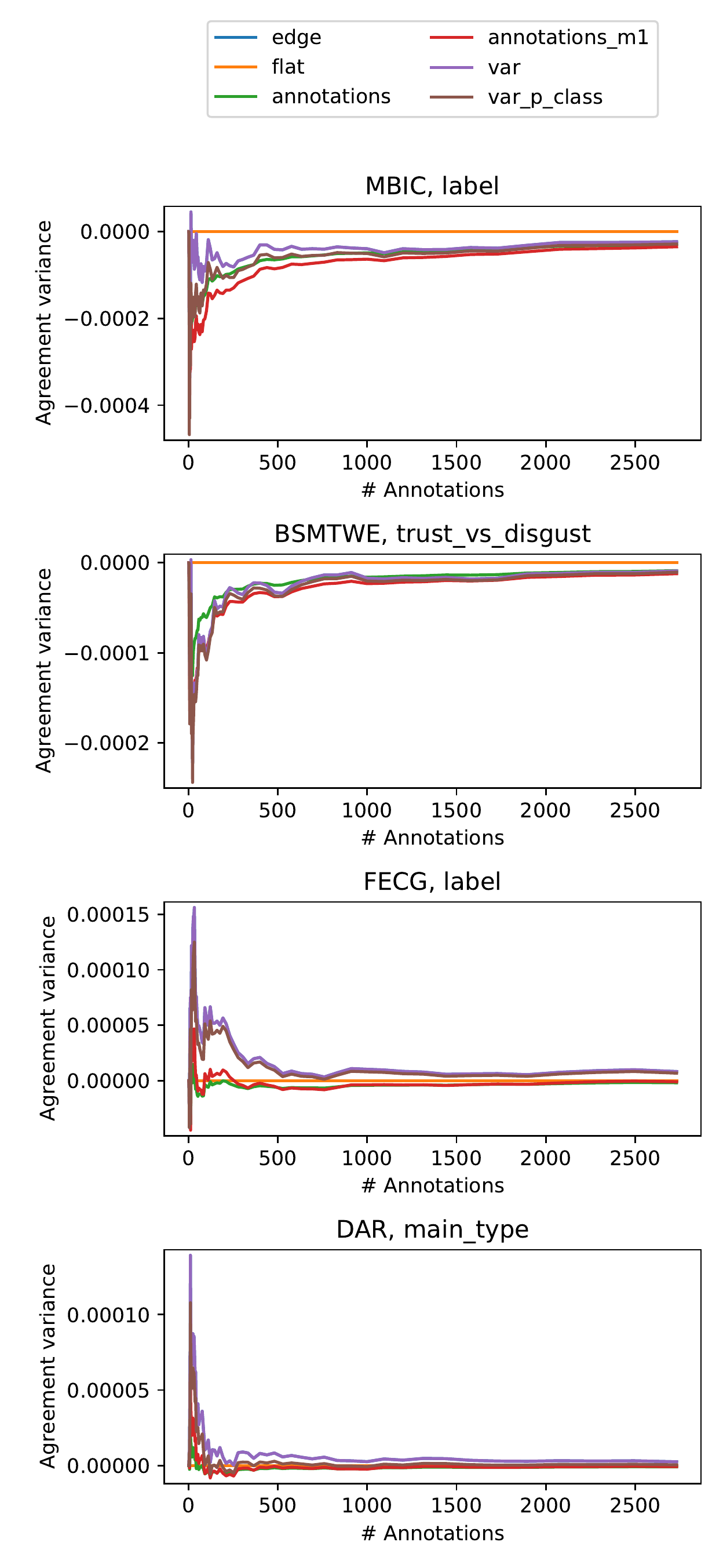}
	\caption{Variance of SPA using different weighing schemes, subtracted by the variance of using \texttt{flat}-weights (for comparison). Note  variance decreases with increasing annotations.} 
	\label{fig:weighing_subtract_mean}
\end{figure}

\subsection{Datasets} \label{sec:datasets}

The following datasets have fully published annotation data, and are used in the experiments. A more detailed description can be found in appendix \ref{app:datasets}.
\begin{description}
	\item[\texttt{[DAR]}] \textbf{Discourse Acts on Reddit} \\
		\citet{zhang2017characterizing} \\
		A corpus and discourse annotations on approximately 115.000 posts from Reddit labelled into 10 classes by 1-3 annotators each.
	\item[\texttt{[MBIC]}] \textbf{A Media Bias Annotation Dataset Including Annotator Characteristics} \\
		\citet{Spinde2021MBIC} \\
		1700 statements representing various media bias instances, labelled as Biased or Non-biased by 9-12 annotators each.
	\item[\texttt{[BSMTWE]}] \textbf{Brazilian Stock Market Tweets with Emotions} \\
		\citet{vieira2020stock} \\
		4553 samples comprising tweets from the Brazilian stock-market domain as Distrust, Trust, Don't Know and Neutral by 1-6 annotators each.
	\item[\texttt{[FECG]}] \textbf{Facial Expression Comparison (Google)} \\
		\citet{vemulapalli2019compact} \\
		51,042 face image triplets annotated into three labels by 5-12 annotations each.
\end{description}

\subsubsection{Annotator Agreement}

The \texttt{MBIC} dataset reports Fleiss kappa, but does not detail how exactly they compute this, despite having a varying number of annotations for the samples. The \texttt{DAR} dataset uses Krippendorff's disagreement measurement instead, as it can be used on sparse data. In the paper for the \texttt{BSMTWE} dataset they specifically note that "\textit{annotators ended up annotating different sets, making it impossible to measure inter-annotator agreement}". Finally in the \texttt{FECG} they do not report an overall inter-annotator agreement, but rather report the number of samples with "strongly agreeing" annotators (two-thirds majority) and number of samples with "weakly agreeing" annotators (unique majority class), as well as the total number of samples. 
These papers highlight the lack of a proper measurement of agreement on datasets with sparse annotations.

\subsection{Increasing Number of Annotated Samples}

Say we start out with zero annotations for a dataset. We now include more and more samples, with an equal number of annotations-per-sample, to the dataset. As the annotations-per-sample is constant, we do not need weights $\vk$ to compute SPA.
We simulate this scenario by randomly subsampling annotations of the above datasets.
We perform 3000 such random subsampling rounds and compute the variance of the resulting SPA. 
Furthermore we test with varying number of annotations-per-sample, depending on what is available in the datasets.

We conduct a similar experiment where we randomly add single annotations, so that samples will have a varying number of annotations. Samples with one or zero annotations are always disregarded, but the remaining samples are used to compute SPA with the \texttt{flat}-weights. In Figure~\ref{fig:subsampling_annotations_sparse_comparison} we plot the variances of SPA with constant annotations-per-sample (coloured lines), together with the variance of SPA when having a varying annotations-per-sample. The x-axis show the number of \textit{annotations} (disregarding 1-annotation samples), in order for the schemes to be comparable. The sparse inter-annotator agreement has similar variance to the 2-annotation curve, which seems reasonable, as randomly adding annotations to these large datasets will create way more 2-annotation samples that multi-annotation samples. As expected, the variance of the estimate decreases with more annotations, as well as with more annotations-per-sample.

\subsection{SPA Weighing}

We compute the variance-over-annotations for the different weighing schemes. We subtract the variance of SPA using \texttt{flat}-weights (baseline) from these curves and plot then in Figure~\ref{fig:weighing_subtract_mean}. We also compute the sum-under-curve for a qualitative comparison in Table~\ref{tab:weights_comparison_table}.
While the weighing schemes in general have relatively little effect on the variance, \texttt{annotations\_m1} works well for all four datasets and is also straightforward to compute.

\section{Conclusion}

Inter-annotator agreement is hard to determine when not all annotators have annotated all instances. This paper presents an agreement measure, Sparse Probability of Agreement (SPA), which can accurately measure inter-annotator agreement without having all annotator-sample pairs available. 

We theoretically show that this is an unbiased estimator for the true Probability of Agreement, and further show that estimate variance can be reduced using weighing schemes. Empirical results over a range of datasets show that SPA is a good estimator of annotation accuracy. We also describe five sample weighing schemes for enhancing annotation assessments, and find that our \texttt{annotations\_m1} weighing scheme can improve annotation agreement accuracy. 

\section{Limitations}
While the presented annotation agreement measure SPA offers an improved signal of annotation quality, the improvement is only offered in the common case of different annotators contributing to different subsets of a dataset. The measure doesn't give greater fidelity if every annotator has labelled every data instance.

\bibliographystyle{acl_natbib}
\bibliography{dataset_refs,refs}

\appendix

\clearpage

\section{Single Item Variance - No Class Distribution}  \label{apdx:item_variance_no_class_distribution}

Consider an item that has been annotated by $n$ annotators into $C$ categories. 
For any category with $\evn_c$ annotations, then number of agreeing annotation-pairs is
\begin{align}
	\frac{\evn_c(\evn_c - 1)}{2}.
\end{align}

We will enumerate all possible combinations of $n$ annotations into $C$ categories, and determine the variance of agreement
\begin{align}
	&\var[\hat{P}]  \label{eq:var_hat_P}
		= \E[\hat{P}^2] - \E[\hat{P}]^2 \\
		&= \frac{1}{|\mathcal{P}^n_C|} \sum_{ \vn \in \mathcal{P}^n_C } 
			\left(\frac{1}{N} \sum_c \frac{\evn_c(\evn_c - 1)}{2}\right)^2
			- \E[\hat{P}]^2 \nonumber
\\
		&= \frac{1}{N^2 C^n} \sum_{ \vn \in \mathcal{P}^n_C } 
			\left(\sum_c \frac{\evn_c(\evn_c - 1)}{2}\right)^2
			- \E[\hat{P}]^2.
\nonumber
\end{align}

Let's consider the squared sum
\begin{align}
	&\left(\sum_c \frac{\evn_c(\evn_c - 1)}{2}\right)^2
\\
		&\;\;\;\;= \sum_{c, c'} \frac{\evn_c(\evn_c - 1)}{2} \frac{\evn_{c'}(\evn_{c'} - 1)}{2} \nonumber
\\
		&\;\;\;\;= \sum_{c, c'} \frac{\evn_c \evn_{c'}(\evn_c - 1)(\evn_{c'} - 1)}{4} \nonumber  \\
		&\;\;\;\;= \frac{1}{4}\sum_{c \not= c'} \evn_c \evn_{c'}(\evn_c - 1)(\evn_{c'} - 1)  \nonumber 
\\
			&\;\;\;\;\;\;\;\;+ \frac{1}{4} \sum_{c} \evn_c^2 (\evn_c - 1)^2
 \nonumber 
\end{align}

\subsection{Self-Pair Sum}
We now denote the last sum as the self-pair sum ($sps(\vn)$):
\begin{align}
	sps(\vn) = \frac{1}{4} \sum_{c} \evn_c^2 (\evn_c - 1)^2.
\end{align}

We can compute the sum of $sps(\vn)$ by considering all possible values for $\evn_c$ and determining how many permutations have this value ($|\{ \evn_c \in  \mathcal{P}^n_C\}|$)
\begin{align}
	\sum_{ \vn \in \mathcal{P}^n_C } sps(\vn) 
		&= \frac{1}{4} \sum_{ \vn \in \mathcal{P}^n_C } \sum_{c} \evn_c^2 (\evn_c - 1)^2
 \\
		&= \frac{1}{4} \sum_{c} |\{ \evn_c \in  \mathcal{P}^n_C\}| \evn_c^2 (\evn_c - 1)^2.
\nonumber
\end{align}

The probability of $n_0 = v$ is (sample from a binomial distribution)
\begin{align}
	p(n_0 = v) = {n \choose v} \left(\frac{1}{C}\right)^v \left( \frac{C - 1}{C} \right)^{n - v}.
\end{align}

The probability of any category getting $v$ annotations is
\begin{align}
	p(\exists &c, \evn_c = v)   \label{eq:prob_with_n_c}
\\
		&= C {n \choose v} \left(\frac{1}{C}\right)^v \left( \frac{C - 1}{C} \right)^{n - v} \nonumber \\
		&= C {n \choose v} C^{-v} (C - 1)^{n - v} C^{v - n} \nonumber \\
		&= {n \choose v} C^{1 - n} (C - 1)^{n - v}. \nonumber
\end{align}

Thus the number of permutations where any category has $\evn_c$ counts is therefore \veri
\begin{align}
	K_{sps}&(\evn_c) = |\{ \evn_c \in  \mathcal{P}^n_C\}| \label{eq:permutations_with_n_c}
\\
		&= |\mathcal{P}^n_C| {n \choose v} C^{1 - n} (C - 1)^{n - v} \nonumber \\
		&= C^n {n \choose v} C^{1 - n} (C - 1)^{n - v} \nonumber \\
		&= {n \choose \evn_c} C (C - 1)^{n-\evn_c}.
\nonumber
\end{align}

So we have \veri
\begin{align}
	&\sum_{ \vn \in \mathcal{P}^n_C } sps(\vn)  \label{eq:sps_sum}
\\
		&= \frac{1}{4} \sum_{ \vn \in \mathcal{P}^n_C } \sum_{c} \evn_c^2 (\evn_c - 1)^2 \nonumber \\
		&= \frac{1}{4} \sum_{\evn_c=0}^n K_{sps}(\evn_c) \; \evn_c^2 (\evn_c - 1)^2. \nonumber
\end{align}

\subsection{Cross-Pair Sum}

The cross-pair sum is
\begin{align}
	cps(\vn) &= \frac{1}{4} \sum_{c \not= c'} \evn_c \evn_{c'}(\evn_c - 1)(\evn_{c'} - 1) 
\end{align}

We wish to compute the sum of $cps(\vn)$ by considering all possible values for $\evn_c$ and factor in the number of permutations
\begin{align}
	\sum_{ \vn \in \mathcal{P}^n_C } cps(\vn) 
		&= \frac{1}{4} \sum_{ \vn \in \mathcal{P}^n_C } \sum_{c \not= c'} \evn_c \evn_{c'}(\evn_c - 1)(\evn_{c'} - 1) 
\nonumber
\end{align}

The probability of selecting the $0$'th category $n_0$ times and the $1$'st category $n_1$ times is\veri
\begin{align}
	p(&n_0, n_1)  \label{eq:prob_n0_n1}
\\
		&= {n \choose n_0} {n - n_0 \choose n_1} \left(\frac{1}{C}\right)^{(n_0 + n_1)} \nonumber \\
			&\qquad \left( \frac{C - 2}{C} \right)^{n - n_0 - n_1} \nonumber \\
		&= {n \choose n_0} {n - n_0 \choose n_1} C^{-(n_0 + n_1)} \nonumber \\
			&\qquad (C - 2)^{n - n_0 - n_1} C^{-(n - n_0 - n_1)} \nonumber \\
		&= {n \choose n_0} {n - n_0 \choose n_1} C^{-n} (C - 2)^{n - n_0 - n_1}. \nonumber
\end{align}

We do not care which two categories are selected, and so we have\veri
\begin{align}
	p(&\exists (c, c'), \evn_c = n_0, \evn_{c'} = n_1)  \label{eq:p_cross_pais_n0_n1}
\\
		&= 2^{\I(n_0 \not= n_1)} C (C - 1) p(n_0, n_1) \nonumber \\
		&= 2^{\I(n_0 \not= n_1)} C (C - 1)
		{n \choose n_0} {n - n_0 \choose n_1} \nonumber \\
			&\qquad C^{-n} (C - 2)^{n - n_0 - n_1} \nonumber \\		
		&= 2^{\I(n_0 \not= n_1)}
		{n \choose n_0} {n - n_0 \choose n_1} \nonumber \\
			&\qquad C^{1-n} (C - 1) (C - 2)^{n - n_0 - n_1}. \nonumber
\end{align}

Thus the number of permutations where any categories have $n_0$ and $n_1$ counts is therefore \veri
\begin{align}
	  |\{ (n_0, n_1)& \in \mathcal{P}^n_C\}| \label{eq:cross_pais_n0_n1}
\\
		&= |\mathcal{P}^n_C| 2^{\I(n_0 \not= n_1)}
			{n \choose n_0} {n - n_0 \choose n_1} C^{1-n} \nonumber \\
			&\qquad (C - 1) (C - 2)^{n - n_0 - n_1}\nonumber \\
		&= C^n 2^{\I(n_0 \not= n_1)}
			{n \choose n_0} {n - n_0 \choose n_1} C^{1-n} \nonumber \\
			&\qquad (C - 1) (C - 2)^{n - n_0 - n_1}\nonumber \\
		&= 2^{\I(n_0 \not= n_1)}
			{n \choose n_0} {n - n_0 \choose n_1} C (C - 1) \nonumber \\
			&\qquad (C - 2)^{n - n_0 - n_1}.\nonumber
\end{align}

We define the following function
\begin{align}
	K_{csp}&(n_0, n_1) = |\{ (n_0, n_1) \in \mathcal{P}^n_C\}| 2^{-\I(n_0 \not= n_1)} \nonumber \\
		&= {n \choose n_0} {n - n_0 \choose n_1} C (C - 1) \nonumber \\
				&\qquad (C - 2)^{n - n_0 - n_1}.
\end{align}

\breakcolumn
So we have\veri
\begin{align}
	&\sum_{ \vn \in \mathcal{P}^n_C } cps(\vn)  \label{eq:cps_sum}
\\
		&= \frac{1}{4} \sum_{ \vn \in \mathcal{P}^n_C } \sum_{c \not= c'} n_0 n_1(n_0 - 1)(n_1 - 1) \nonumber \\
		&= \frac{1}{4} \sum_{n_0=0}^n \sum_{n_1=0}^{\min(n_0, n - n_0)} 2^{\I(n_0 \not= n_1)} \\
			&\qquad K_{csp}(n_0, n_1) \; n_0 n_1(n_0 - 1)(n_1 - 1) \nonumber \\
\end{align}

\subsection{Variance}

We can now compute the variance of an annotated sample of all possible annotation-combinations in $O(n^2)$ time as
\begin{align}
	\var[&\hat{P}]
	= \frac{1}{4 N^2 C^n} \bigg(\sum_{n_0=0}^n \sum_{n_1=0}^{\substack{\min \\ n_0, n - n_0}} 2^{\I(n_0 \not= n_1)} \nonumber \\
		& K_{csp}(n_0, n_1) n_0 n_1(n_0 - 1)(n_1 - 1) \nonumber \\
		&+ \sum_{\evn_c=0}^n K_{sps}(\evn_c) \; \evn_c^2 (\evn_c - 1)^2 \bigg) \nonumber \\
		&- \E[\hat{P}]^2. \label{append:eq:variance_no_class_distribution}
\end{align}

\clearpage

\section{Single Item Variance - Class distribution}
 \label{apdx:item_variance_w_class_distribution}

\subsection{Variance}

We will enumerate all possible combinations of $n$ annotations into $C$ categories, and determine the variance of agreement
\begin{align}
	\var&[\hat{P}]  \label{eq:C_dist:var_hat_P}
		= \E[\hat{P}^2] - \E[\hat{P}]^2 \\
		&= \sum_{ \vn \in \mathcal{P}^n_C } 
			\left(\frac{1}{N} \sum_c \frac{\evn_c(\evn_c - 1)}{2}\right)^2 p(\vn)
\nonumber \\
			&\qquad- \E[\hat{P}]^2
\nonumber \\
		&= \frac{1}{N^2} \sum_{ \vn \in \mathcal{P}^n_C } 
			\left(\sum_c \frac{\evn_c(\evn_c - 1)}{2}\right)^2 p(\vn)
\nonumber \\
			&\qquad- \E[\hat{P}]^2.
\nonumber
\end{align}

Consider the squared sum
\begin{align}
	&\left(\sum_c \frac{\evn_c(\evn_c - 1)}{2}\right)^2
\\
		&\quad= \frac{1}{4}\sum_{c \not= c'} \evn_c \evn_{c'} (\evn_c - 1)(\evn_{c'} - 1) \nonumber \\
		&\quad\quad+ \frac{1}{4} \sum_{c} \evn_c^2 (\evn_c - 1)^2.
\nonumber
\end{align}

\subsection{Self-Pair Sum}
We now denote the last sum as the self-pair sum ($sps(\vn)$):
\begin{align}
	sps(\vn) = \frac{1}{4} \sum_{c} \evn_c^2 (\evn_c - 1)^2.
\end{align}

The probability of $\evn_c$ is (sample from a binomial distribution)
\begin{align}
	p(\evn_c) = {n \choose \evn_c} p_c^{\evn_c} (1 - p_c)^{n - \evn_c}.
\end{align}

So we can compute the self-pair sums by \veri
\begin{align}
	\sum_{ \vn \in \mathcal{P}^n_C } & p(\vn) sps(\vn) \label{eq:C_dist:sps_sum}
\\
		&= \frac{1}{4} \sum_{c} \sum_{\evn_c}^n p(\evn_c) \evn_c^2 (\evn_c - 1)^2. \nonumber
		%
\end{align}

\subsection{Cross-Pair Sum}

The cross-pair sum is
\begin{align}
	cps(\vn) &= \frac{1}{4} \sum_{c \not= c'} \evn_c \evn_{c'}(\evn_c - 1)(\evn_{c'} - 1) 
\end{align}

The probability of $\evn_c$ and $\evn_{c'}$ is
\begin{align}
	p(\evn_c, \evn_{c'})  \label{eq:C_dist:prob_n0_n1}
		&= {n \choose \evn_c} {n - \evn_c \choose \evn_{c'}} \\ 
		&\qquad p_c^{\evn_c} p_{c'}^{\evn_{c'}} (1 - p_c - p_{c'})^{n - \evn_c - \evn_{c'}}.
\nonumber
\end{align}

We can therefore compute the cross-pair sums by \veri
\begin{align}
	&\sum_{ \vn \in \mathcal{P}^n_C } p(\vn) cps(\vn)  \label{eq:C_dist:cps_sum}
 \\
		&= \frac{1}{4} \sum_{ \vn \in \mathcal{P}^n_C } p(\vn) \sum_{\substack{c, c' \\ c \not= c'}} \evn_c \evn_{c'}(\evn_c - 1)(\evn_{c'} - 1) \nonumber \\
		&= \frac{1}{4} 
			\sum_{\substack{c}} \sum_{\substack{c' \\ c \not= c'}} 
			\sum_{\evn_c}^n \sum_{\evn_{c'}}^{n - \evn_c} \nonumber \\ 
			&\qquad p(\evn_c, \evn_{c'}) \evn_c \evn_{c'}(\evn_c - 1)(\evn_{c'} - 1). \nonumber
\end{align}

\subsection{Variance}

We can now compute the variance of an annotated sample using the class distribution in $O(n^2C^2)$ time by \veri
\begin{align}
	&\var[\hat{P}]
	= 
		\frac{1}{4N^2} 
		\sum_{\substack{c}} \sum_{\substack{c' \\ c \not= c'}} 
		\sum_{\evn_c}^n \sum_{\evn_{c'}}^{n - \evn_c} \label{append:eq:variance_with_class_distribution} \\ 
			&\qquad p(\evn_c, \evn_{c'}) \evn_c \evn_{c'}(\evn_c - 1)(\evn_{c'} - 1) \nonumber \\
		&+ \frac{1}{4N^2} \sum_{c} \sum_{\evn_c}^n p(\evn_c) \evn_c^2 (\evn_c - 1)^2 \nonumber
\\
		&- \E[\hat{P}]^2.
\nonumber
\end{align}

\breakcolumn
\subsection{Special Case: Uniform Class Distribution}  \label{append:var_with_class_distribution_uniform_special_case}

In the case of uniform class distribution we have
\begin{align}
	p_c &= C^{-1} \\
	1 - p_c &= (C - 1) C^{-1} \nonumber \\
	1 - p_c - p_{c'} &= (C - 2) C^{-1}. \nonumber
\end{align}

The two probability terms therefore becomes
\begin{align}
	p(\evn_c) 
		&= {n \choose \evn_c} C^{-\evn_c} ((C - 1) C^{-1})^{n - \evn_c} \\
		&= {n \choose \evn_c} C^{-\evn_c} (C - 1)^{n - \evn_c} C^{\evn_c - n} \nonumber \\
		&= {n \choose \evn_c} C^{-n} (C - 1)^{n - \evn_c} \nonumber \\
		&= C^{-n - 1} K_{sps}(\evn_c), \nonumber
\end{align}%
\begin{align}
	p&(\evn_c, \evn_{c'})
		= {n \choose \evn_c} {n - \evn_c \choose \evn_{c'}} \\ 
		&\qquad C^{-\evn_c} C^{-\evn_{c'}} ((C - 2) C^{-1})^{n - \evn_c - \evn_{c'}} \nonumber \\
		&= {n \choose \evn_c} {n - \evn_c \choose \evn_{c'}} \nonumber \\ 
		&\qquad C^{-\evn_c -\evn_{c'}} (C - 2)^{n - \evn_c - \evn_{c'}} C^{\evn_c + \evn_{c'} - n} \nonumber \\
		&= {n \choose \evn_c} {n - \evn_c \choose \evn_{c'}} (C - 2)^{n - \evn_c - \evn_{c'}} C^{- n} \nonumber \\
		&= C^{-n - 1} (C - 1)^{-1} K_{csp}(\evn_c, \evn_{c'}). \nonumber
\end{align}

So the variance becomes
\begin{align}
	&\var[\hat{P}]
	= 
		\frac{1}{4N^2C^{n+1} (C - 1)}
		\sum_{\substack{c}} \sum_{\substack{c' \\ c \not= c'}} 
		\sum_{\evn_c}^n \sum_{\evn_{c'}}^{n - \evn_c} \nonumber \\ 
			& K_{csp}(\evn_c, \evn_{c'}) \evn_c \evn_{c'}(\evn_c - 1)(\evn_{c'} - 1) \nonumber \\
		&+ \frac{1}{4N^2C^{n+1}} \sum_{c} \sum_{\evn_c}^n K_{sps}(\evn_c) \evn_c^2 (\evn_c - 1)^2 \nonumber \\
		&- \E[\hat{P}]^2 
\end{align}

\breakcolumn

\begin{align}
	&= 
		\frac{1}{4N^2C^{n+1} (C - 1)}
		C (C - 1) 
		\sum_{n_0 = 0}^n \sum_{n_1 = 0}^{n - n_0} \nonumber \\ 
			&\;\; K_{csp}(n_0, n_1) n_0 n_1 (n_0 - 1)(n_1 - 1) \nonumber \\
		&+ \frac{1}{4N^2C^{n+1}} C \sum_{n_0=0}^n K_{sps}(n_0) n_0^2 (n_0 - 1)^2 \nonumber \\
		&- \E[\hat{P}]^2 \nonumber \\
	&= 
		\frac{1}{4N^2C^{n}} \Bigg(
		\sum_{n_0 = 0}^n \sum_{n_1 = 0}^{\substack{\min \\ n_0, n - n_0}} 2^{\I(n_0 \not= n_1)} \nonumber \\ 
			& K_{csp}(n_0, n_1) n_0 n_1 (n_0 - 1)(n_1 - 1) \nonumber \\
		&+ \sum_{n_0=0}^n K_{sps}(n_0) n_0^2 (n_0 - 1)^2 \Bigg) \nonumber \\
		&- \E[\hat{P}]^2,
\end{align}

which matches the expression in (\ref{append:eq:variance_no_class_distribution}). We can therefore conclude that (\ref{append:eq:variance_no_class_distribution}) is the maximum-entropy, special case of (\ref{append:eq:variance_with_class_distribution}), when the class distribution is uniform.

\subsection{Datasets} \label{app:datasets}

\begin{figure}
	\includegraphics[width=\linewidth]{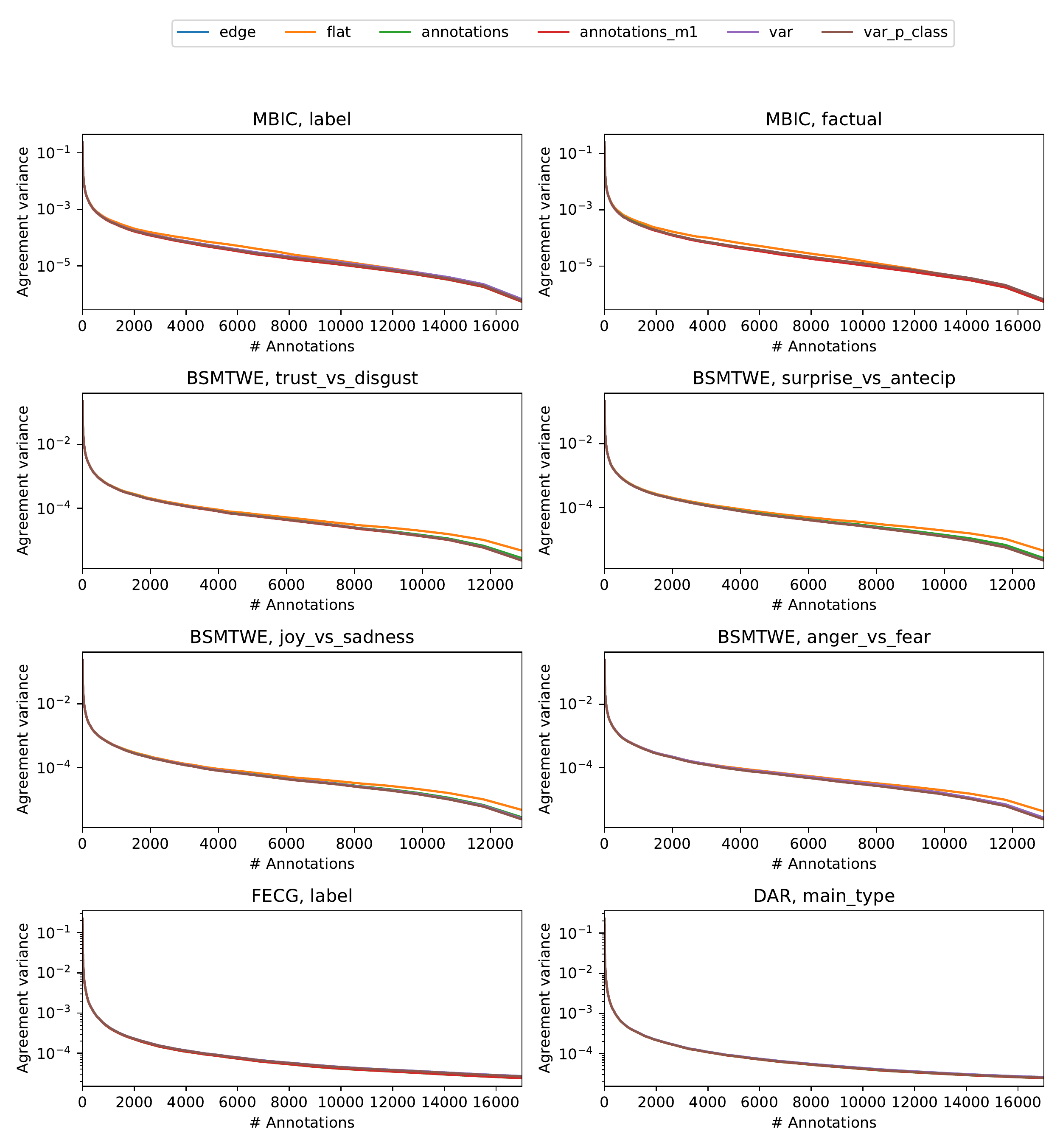}
	\caption{Variance of sparse inter-annotation agreement using different weighing schemes.}
	\label{fig:weighing_full}
\end{figure}

\begin{figure}
	\includegraphics[width=\linewidth]{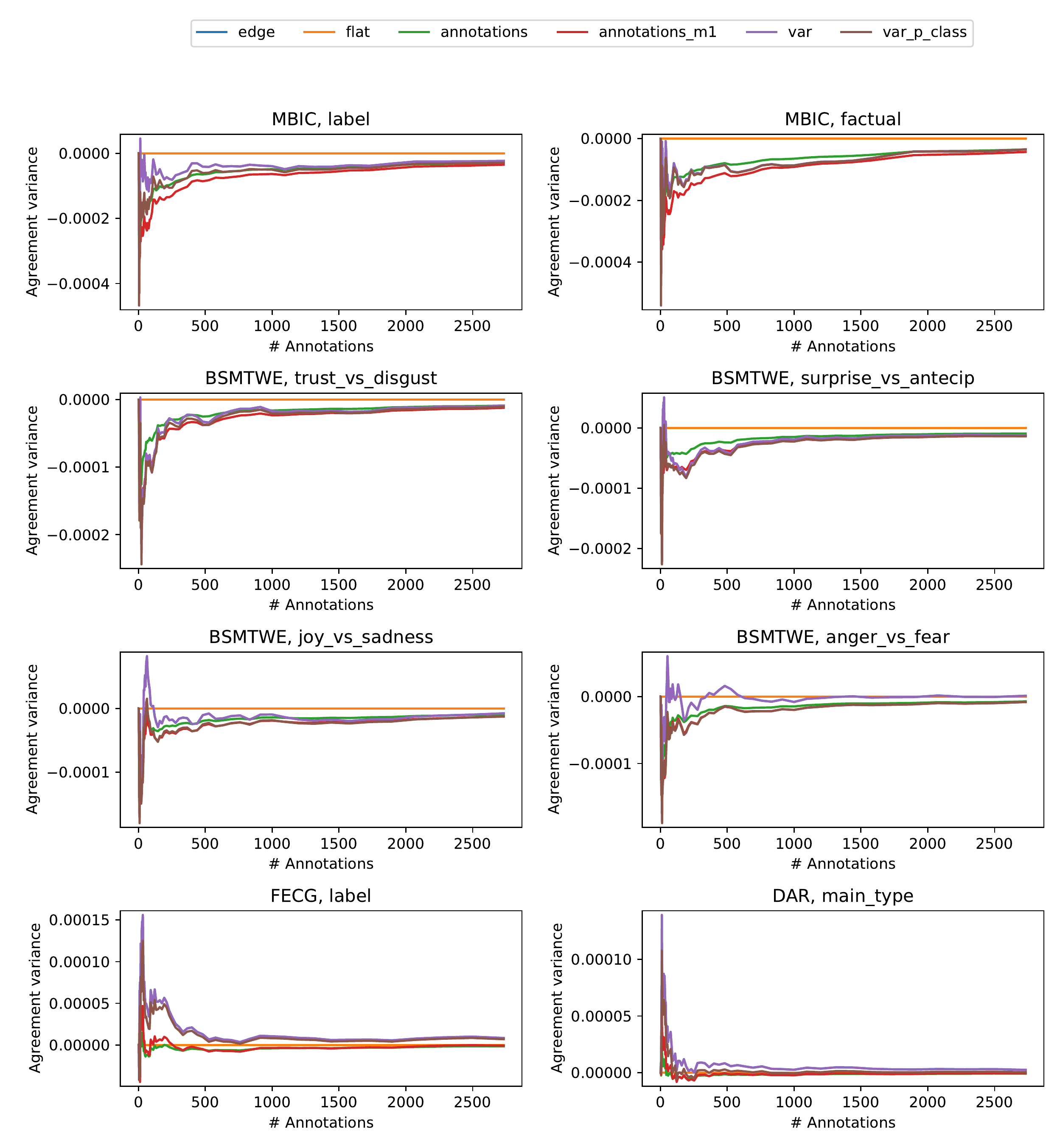}
	\caption{Variance of sparse inter-annotation agreement using different weighing schemes, subtracted by the mean-variance of the methods (for comparison).}
	\label{fig:weighing_subtract_mean_full}
\end{figure}

The following datasets have fully published annotation data, and are used in the experiments. 
In the main paper we only use the first label from each dataset, but in appendix figures \ref{fig:subsampling_samples_full}, \ref{fig:subsampling_annotations_sparse_comparison_full}, \ref{fig:weighing_full} and \ref{fig:weighing_subtract_mean_full}, we show experiments on all labels.

\begin{description}
	\item[\texttt{[DAR]}] \textbf{Discourse Acts on Reddit} \\
		\texttt{DAR} contains a corpus and discourse annotations on approximately 115.000 posts from Reddit. The label used here is the \texttt{main\_type} label, which labels a post as one of 10 classes: 
		\begin{center}
					\begin{tabular}{ll}
					agreement & elaboration \\
					announcement & humor \\
					answer & negative reaction \\
					appreciation & other \\
					disagreement & question
					\end{tabular}
		\end{center}
		\cite{zhang2017characterizing}. Each post is annotated by 1-3 annotators, but some annotators provide multiple labels for a sample. In these cases we randomly select a label, which would decrease the agreement, but make it well-defined for our experiments. 
	\item[\texttt{[MBIC]}] \textbf{A Media Bias Annotation Dataset Including Annotator Characteristics} \\
		\texttt{MBIC} contains 1700 statements representing various media bias instances \cite{Spinde2021MBIC}. The statements are assigned two labels with the following classes
		\begin{description}[font=\normalfont\ttfamily]
			\item[label:] Biased and Non-biased
			\item[factual:] 'Entirely factual', 'Expresses writer’s opinion' and 'Somewhat factual but also opinionated'
		\end{description}
		Each samples is annotated 9-12 times. We use \texttt{label} in the main article but show results on \texttt{factual} in the appendix as well.
	\item[\texttt{[BSMTWE]}] \textbf{Brazilian Stock Market Tweets with Emotions} \\
		\texttt{BSMTWE} contains 4553 samples comprising tweets from the Brazilian stock-market domain, with annotations \cite{vieira2020stock}. There are 4 labels with the following classes
			\begin{description}[font=\normalfont\ttfamily]
				\item[trust\_vs\_disgust:] {DIS}, {TRU}, {dontknow} and {neutral}
				\item[surprise\_vs\_antecip:] {ANT}, {SUR}, {dontknow} and {neutral}
				\item[joy\_vs\_sadness:] {JOY}, {SAD}, {dontknow} and {neutral}
				\item[anger\_vs\_fear:] {ANG}, {FEA}, {dontknow} and {neutral}
			\end{description}
			Each samples is annotated 1-6 times. We use \texttt{trust\_vs\_disgust} in the main article but show results on the other labels in the appendix as well.
	\item[\texttt{[FECG]}] \textbf{Facial Expression Comparison (Google)} \\
		\texttt{FECG} contains 51,042 face image triplets with human annotations, specifying which two faces form the most similar pair in each triplet \cite{vemulapalli2019compact}. There are therefore 3 labels, specifying each of the three pairs in a triplet. We found 5-12 annotations per sample, although almost all samples (50,992) had 6 annotation.
\end{description}

\clearpage

\onecolumn
\section{All Labels All Datasets}

\begin{figure}[ht]
	\includegraphics[width=\linewidth]{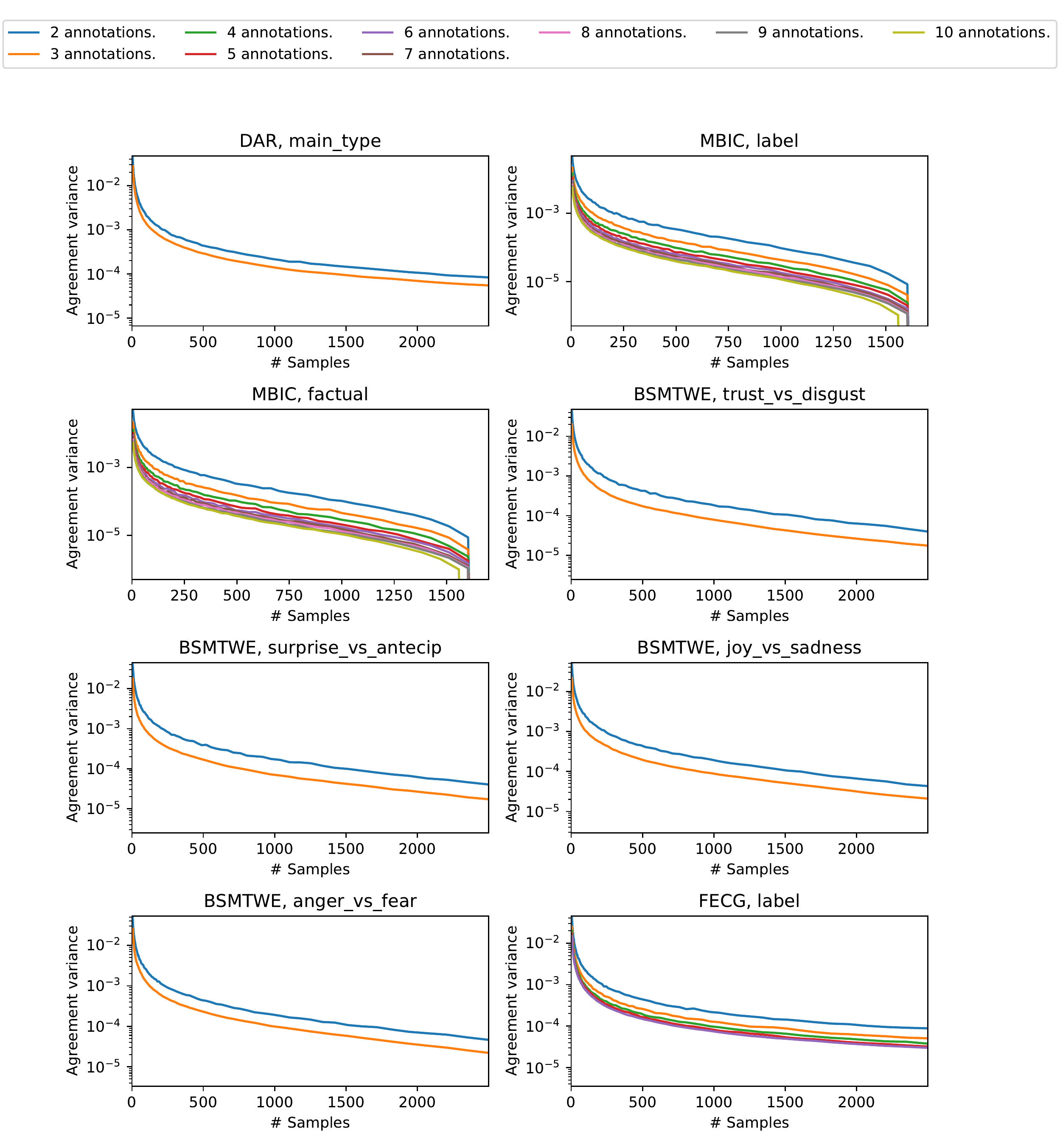}
	\caption{Adding samples to inter-annotator dataset. As more annotated samples are added, the variance in resulting inter-annotator agreement decreases. Note that the y-axes are log-scale.}
	\label{fig:subsampling_samples_full}
\end{figure}

\begin{figure}
	\includegraphics[width=\linewidth]{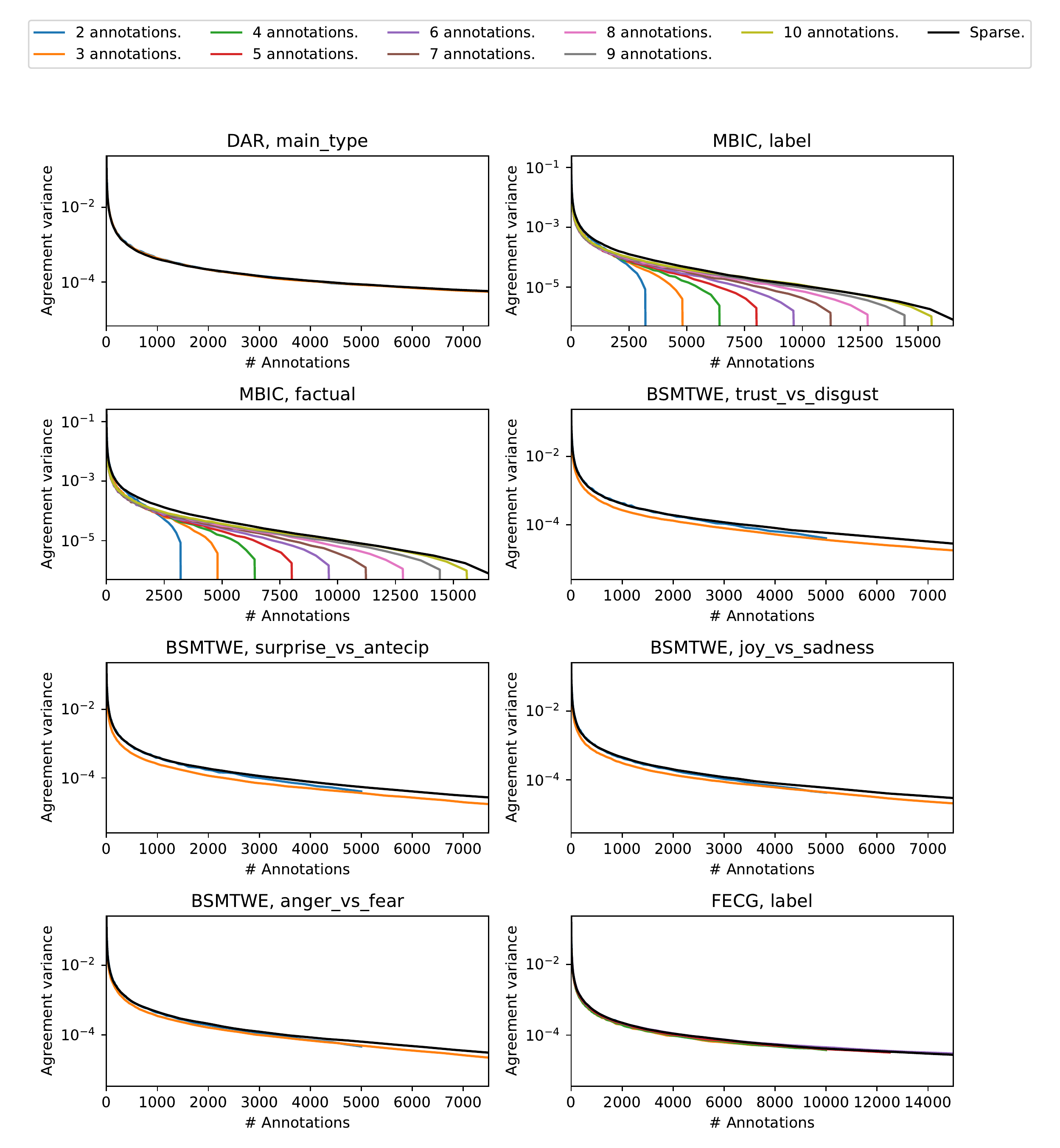}
	\caption{Adding annotations to inter-annotator dataset. As more annotations are added, the variance in resulting inter-annotator agreement decreases. Note that the y-axes are log-scale.}
	\label{fig:subsampling_annotations_sparse_comparison_full}
\end{figure}

\twocolumn





\end{document}